\documentclass{article} % For LaTeX2e
\usepackage{iclr2026_conference,times}
\usepackage[T1]{fontenc}

\usepackage{amsmath,amsfonts,bm}

\def\eqref#1{equation~\ref{#1}}
\def\1{\bm{1}}

\DeclareMathAlphabet{\mathsfit}{\encodingdefault}{\sfdefault}{m}{sl}
\SetMathAlphabet{\mathsfit}{bold}{\encodingdefault}{\sfdefault}{bx}{n}

\usepackage{hyperref}
\usepackage{url}

\usepackage{amsmath}
\usepackage{booktabs}
\usepackage{graphicx}
\usepackage{xspace}
\usepackage{textcomp}

\usepackage{xcolor}
\definecolor{eurekapurple}{RGB}{97,60,211}
\definecolor{citegreen}{RGB}{0,128,0}

\newcommand{\tablecite}[1]{%
  {\footnotesize[%
  \textcolor{citegreen}{\citenum{#1}}%
  ]}%
}
\newcommand{\score}[1]{#1}

\newcommand{\eurekascore}[2]{%
  \if\relax\detokenize{#1}\relax
    \underline{\textcolor{eurekapurple}{\textbf{#2}}}%
  \else
    #1\underline{\textcolor{eurekapurple}{\textbf{#2}}}%
  \fi
}

\newcommand{\method}{\textsc{EurekAgent}\xspace} 

\let\LARGE\Large

\title{%
  \raisebox{-0.1\height}{\includegraphics[height=3.8ex]{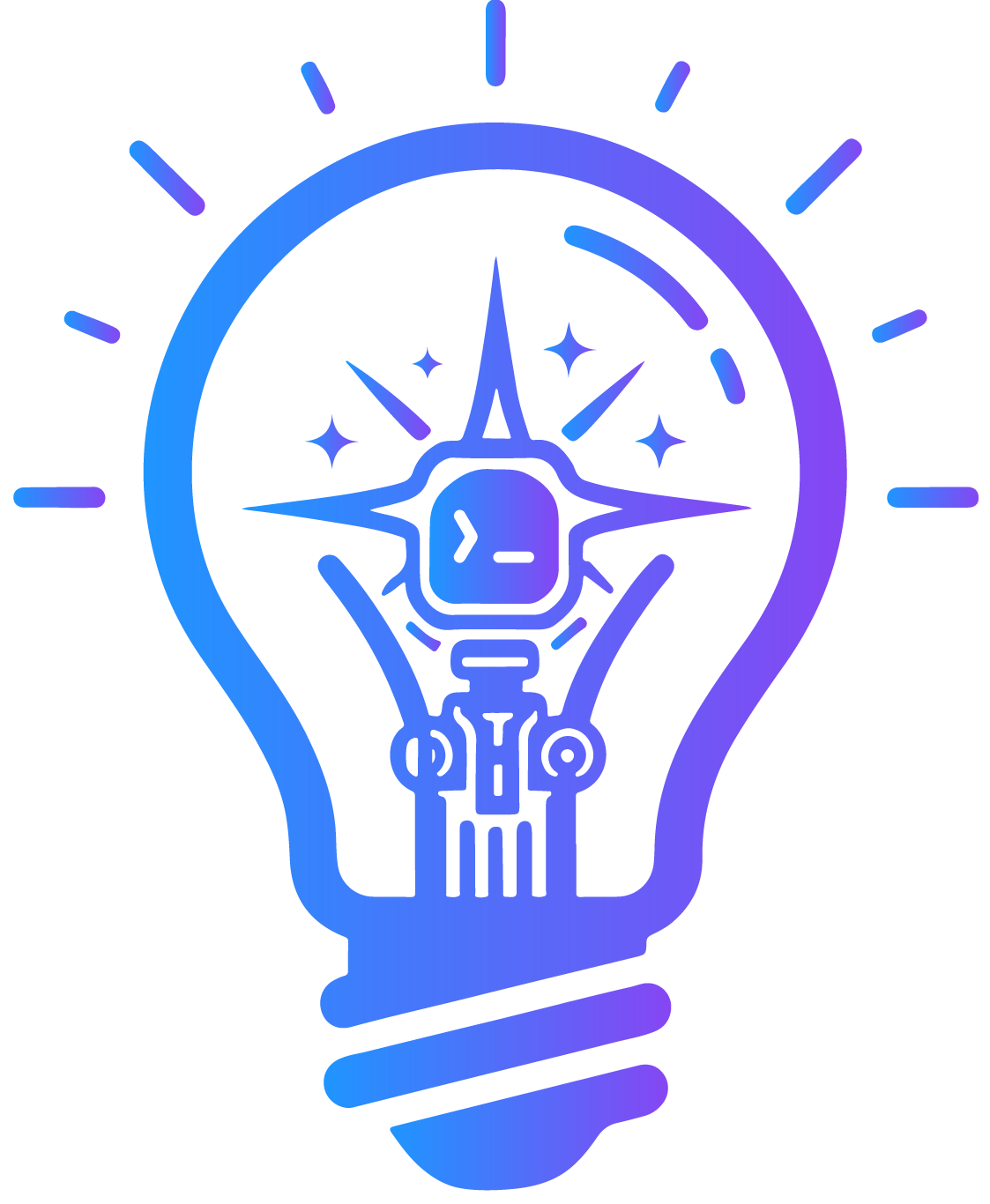}}%
  \hspace{0.4em}%
  \method: Agent Environment Engineering is \\ 
  All You Need for Autonomous Scientific Discovery%
}

\vspace{-0.05in}

\author{%
  \makebox[\textwidth][c]{%
    \begin{tabular}[t]{c}
      Amy Xin$^{1}$, Jiening Siow$^{1}$, Junjie Wang$^{1}$, Zijun Yao$^{1}$, \\
      \textbf{Fanjin Zhang$^{2}$, Jian Song$^{1}$, Lei Hou$^{1}$, Juanzi Li$^{1}$} \\
      {\mdseries $^{1}$Department of Computer Science and Technology, Tsinghua University } \\
      {\mdseries $^{2}$School of Information, Renmin University of China } \\
      {\mdseries\texttt{\{xin-x25, xiaojn25\}@mails.tsinghua.edu.cn}}
    \end{tabular}%
  }%
}

\iclrfinalcopy % Uncomment for camera-ready version, but NOT for submission.
\begin{document}

\maketitle
\lhead{Preprint}
\vspace{-0.15in}

\begin{abstract}

% AI agents 
LLM-based agents 
have shown increasing potential in automating scientific discovery.
% Given a research task and a verifiable objective, they can propose and iterate solutions autonomously well-like a human researcher, and even achieve results that outbeat humans.
% Given a research task and a verifiable objective, they can autonomously propose, evaluate, and iterate solutions, achieving results that even outbeat humans.
% Given verifiable objectives, 
Given an optimizable metric and an execution environment,
% they can autonomously
they can propose, validate, and iterate scientific solutions, 
% sometimes surpassing human-designed results.
% and have produced results that surpass humans.
and have produced results that outperform human-designed approaches.
As model capabilities continue to improve,
we argue that the bottleneck for autonomous scientific discovery is shifting from prescribing agent workflows to designing agent environments---the resources, constraints, and interfaces that shape agent behavior.
% provided to the agent and human supervisor.
% that enable productive exploration while preserving rigor, reliability, and reproducibility.
%  that enable open-ended exploration while preserving rigor, reliability, and reproducibility.
We frame this 
% perspective 
as \textbf{\textit{environment engineering}}: building environments that amplify productive behaviors, such as open-ended exploration, systematic artifact management, and inter-agent collaboration, while suppressing harmful behaviors, such as reward hacking and high-friction human oversight.
% instead of scripting how agents should act through hand-designed workflows, we engineer the environment in which they can safely and productively explore.
% More specifically, we organize environment engineering into four facets: (1) permissions engineering, which exposes useful tools while preventing reward hacking; (2) artifact engineering, which structures code, logs, evaluation results, and shared research state; (3) human-in-the-loop engineering, which enables easy human oversight and intervention; and (4) budget engineering, which bounds time, compute, and token expenditure.
% Building on this principle, 
We present \textbf{\method}, an environment-engineered agent system for 
metric-driven autonomous scientific discovery.
% autonomous discovery 
% % on verifiable research tasks.
% on metric-driven research tasks.
% \method implements environment engineering through four dimensions: permissions engineering, artifact engineering, budget engineering, and human-in-the-loop engineering.
% It coordinates off-the-shelf CLI agents with permission-bounded execution, isolated evaluation, filesystem and git-based collaboration, budget-aware exploration, and easy human-in-the-loop supervision.
% \method engineers the environment along four dimensions: permissions engineering, enforcing permissions-bounded agent execution and isolated evaluation; artifact engineering, enabling filesystem and git-based inter-agent collaboration; budget engineering, allowing budget-aware agent exploration; and human-in-the-loop engineering, supporting easy human supervision and intervention.
\method engineers the environment along four dimensions: permissions engineering for bounded agent execution and isolated evaluation; artifact engineering for filesystem and Git-based collaboration; budget engineering for budget-aware exploration; and human-in-the-loop engineering for easy human supervision and intervention.
% uses off-the-shelf CLI agents as base agent nodes and provides them with permissions-bounded execution, isolated evaluation, filesystem and git-based collaboration, easy human monitoring and intervention, and budget-aware exploration.
% We evaluate \method on three families of verifiable research tasks: algorithmic optimization, machine learning engineering, and kernel engineering.
% \method achieves new state-of-the-art results on all evaluated algorithmic optimization and kernel engineering tasks, and ranks first on the evaluated MLE-Bench subset.
% \method achieves new state-of-the-art results across algorithmic optimization, machine learning engineering, and kernel engineering tasks; notably, it discovers new state-of-the-art 26-circle packing solutions with less than \$11 in total API cost.
% \method sets new state-of-the-art results on multiple math optimization and kernel engineering tasks, ranks first on the evaluated MLE-Bench subset, and discovers new state-of-the-art 26-circle packing solutions with less than \$11 in total API cost.
\method sets new state-of-the-art results on multiple 
% math optimization, 
mathematics,
kernel engineering, and machine learning 
% engineering 
tasks, including new state-of-the-art 26-circle packing results  discovered with less than \$11 in total API cost.
%  with Claude Code and GLM-5.1.
% Notably, \method empowered by Claude Code and GLM-5.1 discovers new state-of-the-art Circle Packing solutions with less than \$11 in total API cost.
% We fully open-source our code and results, hoping to advance environment engineering as a foundation for reliable autonomous research agents.
We open-source our code and results\footnotemark, and call for environment engineering as a core research direction for developing reliable autonomous research agents.

\end{abstract}
% \footnotetext[\dagger]{Corresponding author.}
\footnotetext{\url{https://github.com/THU-Team-Eureka/EurekAgent}}

\begin{figure}[!h]
   \centering
   \includegraphics[width=0.9\linewidth]{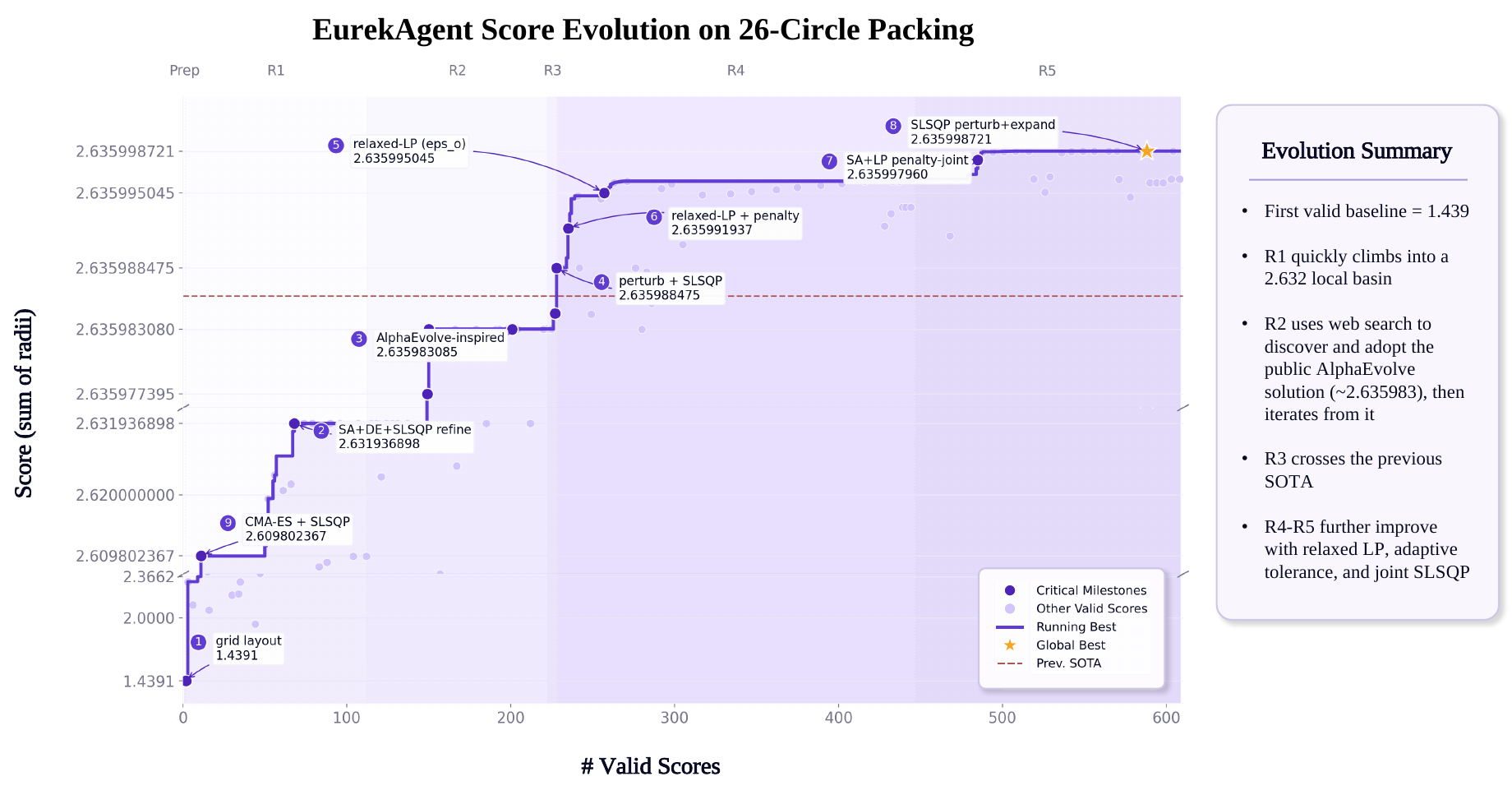}
   % \caption{\method on 26-circle packing: score evolution over the run. The running-best envelope tracks cumulative progress; shaded bands mark preparation and refinement rounds.}
   \caption{\method score evolution progress on the 26-circle packing problem.}
   \label{fig:score-evolution}
   \end{figure}
   
   \begin{table}[!h]
   \centering
   \footnotesize
   \setlength{\tabcolsep}{3pt}
   \renewcommand{\arraystretch}{1.2}
   \resizebox{\linewidth}{!}{%
   \begin{tabular}{@{}l@{\hspace{0.8em}}c@{\hspace{0.9em}}c@{\hspace{0.9em}}c@{\hspace{0.9em}}c@{\hspace{0.9em}}c@{}}
   \toprule
    & \multicolumn{3}{c}{\textbf{Mathematics}} & \multicolumn{1}{c}{\textbf{Kernel Eng.}} & \multicolumn{1}{c}{\textbf{Machine Learning}} \\
   \cmidrule(lr){2-4}\cmidrule(lr){5-5}\cmidrule(lr){6-6}
    & Circle Packing ($\uparrow$) & Erd\H{o}s' Min.\ Overlap ($\downarrow$) & 1st Autocorr. Ineq. ($\downarrow$) & TriMul ($\downarrow$) & MLE-Bench ($\uparrow$) \\
   \midrule
   Prev. Best Human &
     $\sim$ \score{2.634}\tablecite{friedman} &
     \score{0.380927}\tablecite{haugland2016} &
     \score{1.509730}\tablecite{matolcsi2010} &
     \score{\(2096.04\,\mu\mathrm{s}\)} &
     N/A \\
   Prev. Best AI  &
     \score{2.635986}\tablecite{thetaevolve} &
     \score{0.380876}\tablecite{tttdiscover} &
     \score{1.502863}\tablecite{tttdiscover} &
     \score{\(2247.78\,\mu\mathrm{s}\)}\tablecite{tttdiscover} &
     \score{71.43\%}\tablecite{aibuildai} \\
   \midrule
   \method &
     \eurekascore{2.635}{999} &
     \eurekascore{0.3808}{70} &
     \eurekascore{1.5028}{61} &
     \eurekascore{}{2005.03\,\textmu s} &
     \eurekascore{}{85.71\%} \\
   \bottomrule
   \end{tabular}%
   }
   \caption{An overview of \method's performance on metric-driven research tasks across mathematics, kernel engineering, and machine learning. \method sets new state-of-the-art across all mathematics and kernel engineering tasks, and ranks first on the evaluated MLE-Bench subset. ($\uparrow$) denotes higher is better, while ($\downarrow$) denotes lower is better. \emph{Prev.\ Best Human} and \emph{Prev.\ Best AI} denote the best published human and AI results before \method.}
   \label{tab:main-results}
   \end{table}

   \section{Introduction}

   % From research-specific agents such as AlphaEvolve to general-purpose agents such as Claude Code, modern AI agents have showcased strong potential in automating scientific discovery.
   % With the continuous improvement of general and coding capabilities of large language models, LLM-based agents have showcased strong capabilities in automating scientific discovery of verifiable tasks. 
   % Given a verifiable research task and an optimizable metric, AI agents can autonomously iterate solutions with trial-and-error, already able to autonomously discover new state-of-the-art scientific solutions in domains such as mathematics (cite AIDE: AI-Driven Exploration in the Space of Code, alphaevolve, shinkaevolve, thetaevolve, TTT-discover), algorithms and engineering (cite KernelBench, EvoX:  Meta-Evolution for Automated Discovery, Meta-Harness), engineering tasks (cite KernelBench, Meta-Harness) and model training (cite MLE-Bench, AIDE, R\&D Agent, Famou-Agent, AIBuildAI).
   % This shift from manual engineering to computational optimization largely reduces human effort while facilitating exploration in a larger search space, and more efficient trial-and-error / iteration
   Large language models are increasingly transforming scientific discovery from manual trial-and-error to computational exploration:
   in domains where research progress can be measured by
   %  executable artifacts and 
   an optimizable metric, LLM-based agents can autonomously propose hypotheses, run experiments, observe feedback, and iterate solutions, reducing human effort in method tuning while largely expanding the scale of exploration. 
   LLM-based agents have already produced strong results in tasks across domains such as mathematics, algorithms, kernel engineering, and machine learning engineering \citep{aide,alphaevolve,shinkaevolve,thetaevolve,tttdiscover,kernelbench,mlebench,rdagent,aibuildai,li2025fmagent}.
   We envision this as an emerging paradigm shift in scientific research: humans increasingly focus on selecting valuable directions, formulating meaningful metrics, and supervising validity, while agents execute large volumes of methodological exploration.

   Most existing autonomous research systems realize this vision by prescribing research-specific agentic workflows. 
   % A typical system decomposes research into prescribed workflows: idea generation, implementation, execution, reflection, literature or memory retrieval, candidate selection, report writing, and sometimes peer review or debate. 
   Evolutionary systems such as AlphaEvolve explicitly maintain populations of candidate programs and use evaluator feedback to guide mutation and selection \citep{alphaevolve,shinkaevolve,evox}. 
   Machine learning systems such as AIDE organize exploration around solution trees, feedback loops, and role-specialized agents \citep{aide,rdagent}.
   More recent systems introduce structured debate, periodic self-review, and self-learning modules \citep{autoresearchclaw,coral}. 
   % These workflow-based designs have been crucial: they made autonomous research concrete, auditable, and measurable at a time when agents were brittle and needed substantial scaffolding.
   While these designs can be effective, they also encode strong assumptions about how research should proceed. 
   As general-purpose coding agents like Claude Code and Codex become stronger, recent evidence suggests that much of the useful capability may already reside in the base agent: given a clear research task and an optimizable metric, these agents can already discover new state-of-the-art scientific solutions~\citep{autoevolver,karpathy_autoresearch}.
   On ResearchClawBench~\citep{researchclawbench}, a benchmark of 40 research tasks across 10 diverse domains, both Claude Code and Codex, used as standalone general-purpose agents, outperform all evaluated research-specific agent systems.
   
   % First up is reward hacking, already observed in many research scenarios (cite "The More You Automate, the Less You See: Hidden Pitfalls of AI Scientist Systems; GPU Mode's blog "Anatomy of a Reward Hack: A Real Story from the Latest GPU Mode NVFP4 Competition", Opus 4.7 Model System Card)
   However, task performance alone does not make reliable autonomous researchers. 
   Scientific discovery requires rigor, reproducibility, and inspectability, yet agents may contaminate evaluations, manipulate artifacts, or fail to follow procedural constraints. 
   Such reward-hacking and observability failures have already been reported in agentic research systems~\citep{more_you_automate,gpumode_rewardhack,anthropic_opus47_systemcard}. 
   Therefore, trusting agents without environmental constraints can lead to 
   impressive but 
   unreliable results.
   
   % In this paper, we argue that the central bottleneck for autonomous scientific discovery is shifting from \emph{agent behavior scripting} to \emph{agent environment engineering}.
   % This view echoes Gibson's theory of affordances in ecological psychology: an environment shapes the possibilities for actions available to an actor, ``either for good or ill'' \citep{gibson1979ecological}.
   % Instead of hard-coding workflow steps such as reflection, mutation, or synthesis, we should give capable agents an \textit{environment} that affords free exploration, feedback observation, artifact recording, and inter-agent coordination, while imposing responsible permission constraints and providing easy interfaces for human monitoring and intervention. 
   % The analogy is a capable PhD student: productivity comes not from minute-by-minute instructions, but from clear goals, useful resources, 
   % % taste guidance, 
   % accurate feedback, space for trial-and-error, progress records, accountability, and mentor supervision.
   These observations suggest that as general-purpose agents become more capable, the bottleneck for autonomous scientific discovery is shifting from prescribing agent behavior through detailed workflows to engineering the environments in which agents operate.
   We frame this as \emph{environment engineering}.
   This 
   % view 
   echoes Gibson's theory of affordances in ecological psychology: an environment shapes the possibilities for action available to an actor, ``either for good or ill''~\citep{gibson1979ecological}.
   For scientific discovery, a well-engineered environment should suppress harmful affordances such as evaluation tampering and artifact manipulation, while amplifying productive affordances such as free exploration, accurate rewards, inter-agent coordination, and easy human supervision.
   The analogy is a capable PhD student: productivity comes not from minute-by-minute instructions, but from accountability, research autonomy, accurate feedback, peer collaboration, and mentor supervision.
   
   % We present \method, an agent system for autonomous scientific discovery with environment engineering designs.
   % \method coordinates off-the-shelf CLI agents with environment engineering from four facets: (1) permissions engineering, which exposes useful tools while preventing reward hacking; (2) artifact engineering, which structures code, logs, memory, evaluation results, and shared research state; (3) human-in-the-loop engineering, which enables low-friction oversight and intervention; and (4) budget engineering, which bounds time, compute, and token expenditure.
   % Besides this, the agent freely explores within the environment.
   % This design gives agents broad freedom to explore while preserving the integrity, reproducibility, and inspectability required for scientific research.
   We present \method, an agent system for autonomous scientific discovery that coordinates off-the-shelf CLI agents through four environment engineering dimensions: (1) permissions engineering, to expose useful capabilities and resources while preventing research-integrity violations; (2) artifact engineering, to structure solutions, logs, and evaluation results as shared progress memory; (3) budget engineering, to enable budget-aware exploration with runtime and compute boundaries; and (4) human-in-the-loop engineering, to support easy human supervision and intervention.
   % Within this environment, agents are free to explore rather than follow a prescribed workflow.
   Within this environment, the agent remains free to select its own research workflows and strategies. 
   
   % We call for more attention on environment engineering and highlight its importance for developing capable, efficient, and reliable autoresearch agents.
   We evaluate \method on metric-driven research tasks spanning mathematics, kernel engineering, and machine learning engineering. 
   Using off-the-shelf CLI agents and environment-level design, \method achieves new state-of-the-art results across all mathematics and kernel engineering tasks, and ranks first on the evaluated MLE-Bench subset.
   % In particular, \method discovers new best-known solutions on multiple mathematical optimization tasks, including a new state-of-the-art 26-circle packing solution found with less than \$11 total API cost. 
   Furthermore, with Claude Code as the CLI agent and GLM-5.1 as the base model, \method achieves new state-of-the-art results on the three mathematics tasks with an average API cost below \$17, where the 26-circle packing task achieves the lowest API cost of \$11. 
   % These results suggest that carefully engineered environments can unlock capable, efficient, and reliable autonomous scientific discovery. 
   % More broadly,
   We call for environment engineering as a core research direction for building capable, efficient, and responsible autonomous research agents.

   \section{Related Work}
   
   \subsection{Agents for Scientific Discovery}
   
   Autonomous research agents have attracted growing interest 
   % as a way 
   to accelerate scientific with large-scale computational exploration.
   Systems such as The AI Scientist aim to automate scientific research in an end-to-end manner, covering stages such as idea generation, experimentation, and paper
     writing~\citep{aiscientist}.
   Within this broader vision, one especially concrete direction is scientific discovery with verifiable objectives and optimizable metrics, where agents
   autonomously explore and evolve solutions through evaluator feedback.
   In machine learning engineering, systems such as AIDE, R\&D-Agent, AIBuildAI, MLE-STAR, and ML-Master formulate progress as iterative code development guided by
   validation scores~\citep{aide,rdagent,aibuildai,mle_star,mlmaster}.
   In algorithmic and mathematical discovery, training-free solution evolution methods such as FunSearch, AlphaEvolve, ShinkaEvolve, EvoX, AdaEvolve, and OpenEvolve use LLMs to propose or mutate candidate programs under evaluator-guided
   selection~\citep{funsearch,alphaevolve,shinkaevolve,evox,adaevolve,openevolve}.
   More recently, test-time training systems such as ThetaEvolve and TTT-Discover further use the optimizable metric as a reward signal to update the model during 
   exploration~\citep{thetaevolve,tttdiscover}.
   These systems demonstrate the power of evaluator-guided discovery, but they typically use fixed workflows to prescribe core agent behaviors such as proposal, mutation, selection, or reflection.
   \method instead uses strong general-purpose CLI agents as basic nodes, and focuses on engineering an environment that lets agents exercise their own capabilities
   reliability.

   \subsection{Agent Environments and Research Integrity}
   
   As agents become more autonomous, the surrounding environment becomes a central determinant of reliability. 
   % Benchmarks such as AgentBench, MLE-Bench, and KernelBench evaluate agents in executable environments with tool use, code execution, and task-specific scoring~\citep{agentbench,mlebench,kernelbench}.
   % Several recent research-agent systems also begin to introduce environment-level safeguards: MLE-STAR adds leakage checking for machine learning pipelines, AIRA\_2 proposes hidden consistent evaluation, and CORAL hides grader code behind an evaluation interface~\citep{mle_star,aira2,coral}.
   Some recent 
   % autoresearch 
   systems have begun to recognize the importance of environment reliability and introduce safeguards for specific failure modes.
     For example, MLE-STAR adds leakage checking for machine learning pipelines, 
     % AIRA\_2 studies hidden consistent evaluation, 
     and CORAL hides grader code behind an evaluation
     interface~\citep{mle_star,coral}.
   At the same time, analyses of 
   % AI-scientist systems and 
   real reward-hacking incidents show that agents can exploit weak evaluation protocols, contaminate evidence, or violate procedural assumptions~\citep{more_you_automate,gpumode_rewardhack,anthropic_opus47_systemcard}.
   Instruction-following failures in complex agentic settings further suggest that reliability cannot be delegated entirely to prompt engineering~\citep{agentif}.
   Some existing work therefore explores environment design to avoid common failures, but these are usually introduced as task-specific safeguards.
   \method makes environment engineering the central design objective: it organizes permissions, artifacts, budgets, and human oversight as first-class mechanisms for supporting open-ended agent exploration while preserving evaluator integrity, traceability, and reproducibility.
   
   \begin{figure}[!t]
     \centering
     \includegraphics[width=\linewidth]{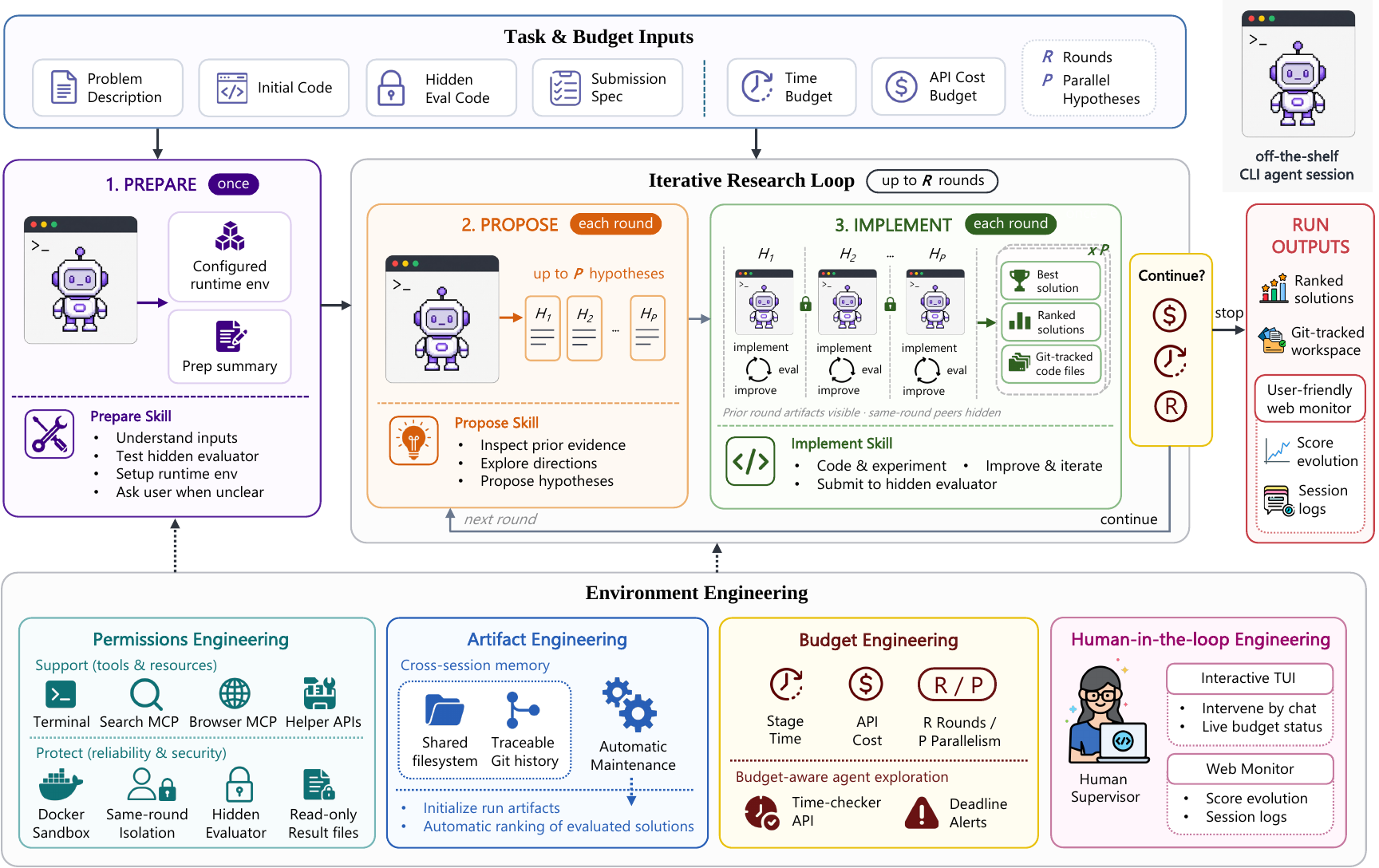}
     \caption{Overview of \method. Given task inputs and budgets, \method executes a prepare stage followed by repeated propose and parallel implement stages, while the environment engineering layer provides secure evaluation, artifact memory, budget control, and human oversight.}
     \label{fig:system-overview}
   \end{figure}

   \section{\method}

   % \method is an environment-engineered agent system for autonomous scientific discovery on metric-driven research tasks.
   % Instead of prescribing detailed research workflows, \method coordinates general-purpose CLI-agent sessions through a structured environment that provides reliable evaluation, shared artifacts, permission boundaries, budget control, and human supervision.
   % In this section, we first describe the overall system loop, and then detail the environment engineering designs that make the loop reliable and inspectable.
   In this section, we present the system design of \method. 
   Figure~\ref{fig:system-overview} summarizes the overall architecture.
   We first overview the overall system loop (~\ref{sec:method_system_overview}), then detail the environment engineering designs
   %  that make the system reliable and inspectable 
    (~\ref{sec:method_env_engineering}).
   
   \subsection{System Overview}
   \label{sec:method_system_overview}
   
   \method is an environment-engineered agent system for metric-driven research tasks. 
   Given a problem description, a hidden evaluation script, a submission-format specification document, optional initial code, and time and API cost budgets, \method 
   % constructs a self-contained research environment and 
   coordinates multiple sessions of off-the-shelf CLI agents 
   % (i.e. Claude Code) 
   to autonomously propose and iterate high-scoring solutions. 
   Instead of prescribing a detailed research workflow,
   \method engineers an outer environment that organizes agent activity through a simple three-stage loop:
     \[
     \textsc{Prepare} \rightarrow
     \left[
     \textsc{Propose}_{r}
     \rightarrow
     \{\textsc{Implement}_{r,p}\}_{p=1}^{P_r}
     \right]_{r=1}^{R}, \quad P_r \le P ,
     \]
     % \[
     %   \textsc{Prepare} \rightarrow
     %   \left[
     %   \textsc{Propose}_{r}
     %   \rightarrow
     %   \underbrace{\{\textsc{Implement}_{r,1},\ldots,\textsc{Implement}_{r,P_r}\}}_{\text{parallel implementation sessions}}
     %   \right]_{r=1}^{R}, \quad P_r \le P .
     % \]
   where \(R\) is the maximum number of 
   % optimization rounds
   iteration rounds
   and \(P\) is the maximum number of parallel implementation sessions per implement stage, both adjustable by the user.
   Each round consists of one proposal session followed by up to \(P\) parallel implementation sessions. 
   Across stages and rounds, the environment only handles outer-loop coordination: it initializes the workspace, transitions between stages, specifies each session's objective and required deliverables, exposes the tool and resource interfaces, records and ranks scored solutions, persists run and session state, and enforces time and cost budgets. Within these boundaries, CLI-agent sessions freely decide their own research strategies, experiment plans, implementation details, and refinement processes.
   % The workspace records public task inputs, per-round state, per-approach work directories, secure grading outputs, code versions, logs, and session transcripts, so the entire discovery process is inspectable, resumable, and reproducible.

   % \paragraph{Prepare Stage.}
   % Ensures clarity for the problem description, verifies the evaluation pipeline, and prepares the environment.
   % (refer to @/Users/ami/Desktop/EurekAgent/ProjectEureka-refined/.claude/skills/prepare-workspace for more info)
   
   % \paragraph{Propose Stage.}
   % For each round, proposes diverse initial hypotheses for the next round.
   % (refer to @/Users/ami/Desktop/EurekAgent/ProjectEureka-refined/.claude/skills/propose-approaches for more info)
   
   % \paragraph{Implement Stage.}
   % For each round, spawns P parallel implementation agents and implements the proposed hypotheses in parallel. The agent can use the hypotheses as a starting point and explore solutions from there.
   % (refer to @/Users/ami/Desktop/EurekAgent/ProjectEureka-refined/.claude/skills/implement-approaches for more info)
   \paragraph{Prepare Stage.}
   Before iteration begins, \method launches a preparation agent session to set up a reliable runtime setup for subsequent solution iteration.
   The agent reads the problem description, the evaluator-facing submission requirements document, and optional initial code; tests the
   hidden evaluation service; and installs or validates required runtime dependencies. If the problem setup is ambiguous or broken, the
   agent can pause and request human clarification rather than allowing optimization to proceed from an unreliable setup. The stage ends
   by writing a preparation summary and a completion artifact, which become shared context for later proposal and implementation sessions.
   This stage is executed only once at the beginning of the research process; after that, \method executes \(R\) rounds of
   propose--implement iteration.
   
   \paragraph{Propose Stage.}
   At the beginning of each iteration round, \method launches a proposal agent session to generate diverse initial hypotheses for the next round of solution
   optimization. The session reads the task inputs, the preparation summary, and the ranked best solutions from previous rounds, if any. It
   may also inspect previous-round workspaces for implementation details and use web search or browsing tools to gather related literature or existing open-source solutions. It then writes a manifest containing up to \(P\) candidate hypotheses
   and creates an implementation-ready description for each hypothesis. This stage acts as the fan-in step of \method: empirical evidence
   from earlier rounds, together with information from the internet, is distilled into a new set of promising, diverse, and independently
   executable research hypotheses.
   
   \paragraph{Implement Stage.}
     The implement stage is the fan-out step of \method. For each proposed hypothesis, \method launches a separate implementation agent session in
     parallel and assigns it a separate workspace. Each session starts from its assigned hypothesis as an initial direction, but may
     iteratively refine, debug, or modify the solution according to feedback from the hidden evaluator. Sessions submit candidate
     solutions through the secure evaluation service, which records all evaluated submissions and maintains both intermediate results and the
     best valid result. After the parallel implement sessions complete or exhaust their budgets, \method automatically ranks all valid submissions, and updates a ranked solution history file as shared context for the next round. This propose-implement loop combines broad parallel exploration with cross-round accumulation of empirical progress,
     continuing for improvement until the budget limits or stage completion conditions are reached.
   
   \subsection{Environment Engineering in \method}
   \label{sec:method_env_engineering}
   
   \method is designed through four environment engineering dimensions: (1) permissions engineering, (2) artifact engineering, (3) budget engineering, and (4) human-in-the-loop engineering.
   % \method implements environment engineering along four dimensions: permissions, artifacts, budgets, and human-in-the-loop supervision.
   % These dimensions are not separate agent roles or hand-written research procedures; rather, they define the external environment in which
   % general-purpose CLI-agent sessions operate. 
   The goal is to grant agent sessions enough affordances to perform open-ended solution optimization,
   while making the research process reliable, inspectable, and resource-bounded.
   
   \begin{figure}[!t]
     \centering
     \includegraphics[width=1\linewidth]{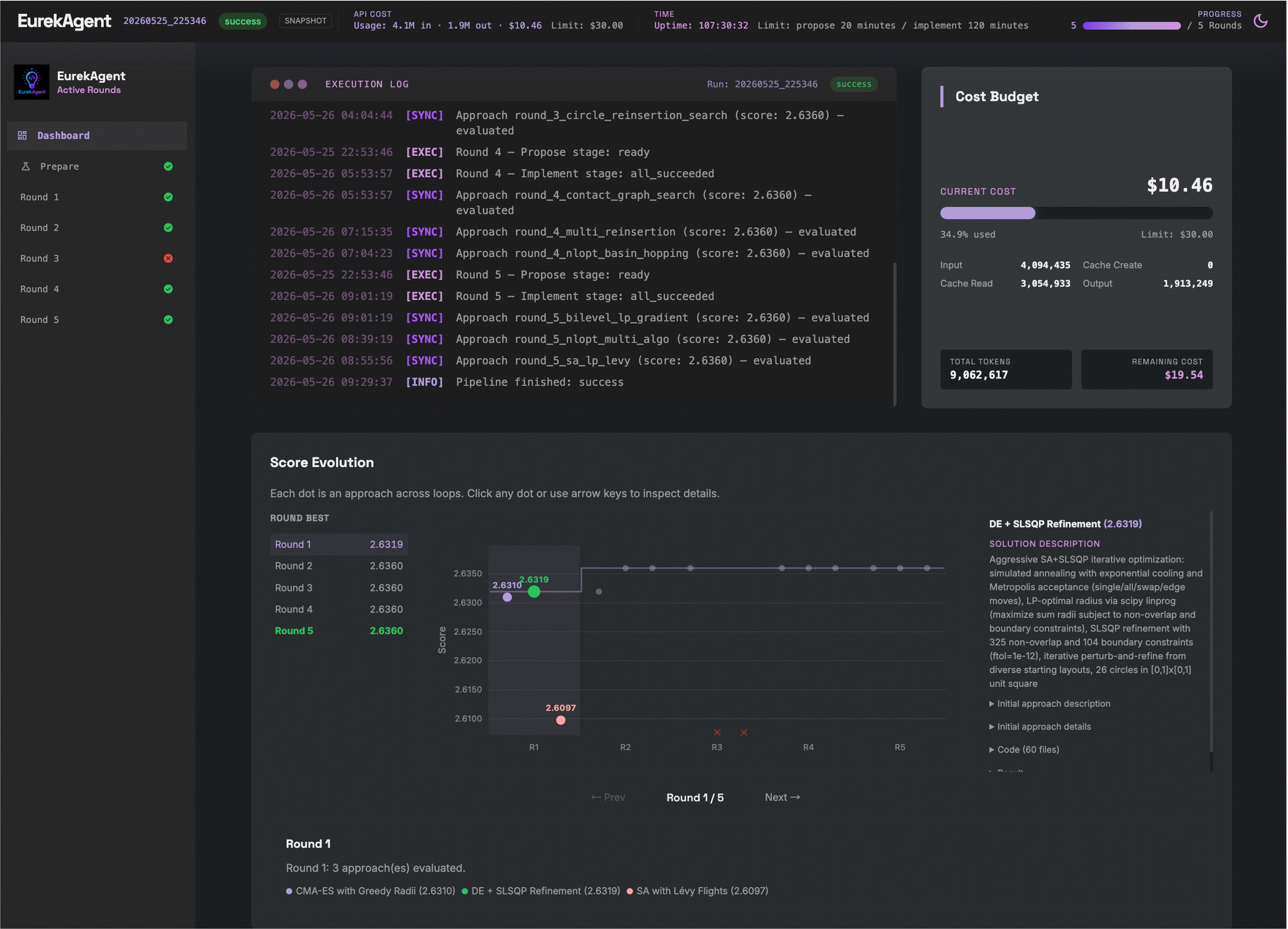}
     \caption{The \method web monitor interface. The monitor provides a user-friendly overview of each run, including status logs, score evolution, per-round and global best approaches, and budget usage. It also records complete session
     transcripts in an organized view, allowing users to inspect full trajectories of every agent session.}
     \label{fig:web-monitor}
     \end{figure}
   
   \paragraph{Permissions Engineering.}
   Scientific discovery agents need broad capabilities, but unconstrained capability can compromise research integrity. 
   % \method provides productive affordances while enforcing integrity constraints via hard permission boundaries. 
   \method imposes system-level permission boundaries to support productive exploration while preventing research-integrity violations. 
   On the productive side, \method provides a freely configurable 
   Python environment, workspace-level shell access, capable web search and browser tools, and full access to the same run's previous-round artifacts.
   This gives agents researcher-like access to tools, files, internet, and prior experience to aid solution iteration. 
   On the constraint side, \method uses run-level isolation and controller-owned interfaces to prevent common failure modes. 
   Each run executes inside a Docker container with a mounted workspace, protecting files
   outside the run from accidental or adversarial modification. 
   The hidden evaluator with possible test data are kept outside the agent-visible workspace and exposed only through a secure grading service: agents can submit candidates and receive official scores, but cannot inspect or modify the
   evaluator itself. 
   The authoritative result files generated by the hidden evaluator
   % including intermediate results and best results, 
   are automatically updated by the system, and hooks are implemented to block agent modification of these controller-owned files. 
   \method also enforces same-round isolation among parallel implementation sessions: an implementation session may learn from previous rounds, but cannot inspect or copy from peer approaches in the same round, reducing premature collapse toward a single local direction. 
   For GPU tasks, \method uses a default-deny policy: GPUs are invisible unless acquired through a provided GPU helper API, which records lock ownership and ensures
   that each physical GPU is held by at most one agent session at a time.
   Together, these mechanisms expose useful resources while removing high-risk affordances such as evaluator leakage, score tampering, uncontrolled GPU contention, and same-round solution copying.
   
   \paragraph{Artifact Engineering.}
   \method uses the filesystem coupled with Git history as shared long-term memory. 
   The filesystem stores stage deliverables for 
   % cross-stage and
   % cross-round communication
   cross-session communication, including preparation summaries, proposal manifests, hypotheses, solution code, evaluator feedback, and scored submissions. 
   \method also maintains system-managed artifacts: web-search history is logged as a cache of explored internet information, and official scores are automatically recorded and ranked.
   %  after each round. 
   The ranked historical solutions enable later agent sessions to quickly identify strong prior solutions and inspect their code, logs, and intermediate results when needed. 
   All run artifacts are persisted under the run directory, providing the persistent substrate for traceability, interruption recovery, and resumability. 
   Within each session, Git commits track solution evolution. 
   We instruct agents to describe both the current standalone solution and what changed from the previous version in each commit message.
   
   \paragraph{Budget Engineering.}
     Autonomous research agents can consume substantial time, compute, and API budget, so \method makes budget limits part of the environment
     settings.
     \method controls resources along two axes: wall-clock time and API cost.
     For time, users specify separate limits for proposal and implementation sessions, reflecting that hypothesis generation and long-running
     solution iteration require different time scales.
     % The appropriate limits are task-dependent, so \method exposes them as user-configurable run parameters rather than fixed constants.
     Furthermore, \method makes agents time-aware through both active and passive mechanisms:
     (i) actively, agents can call a provided time-checking helper API to inspect elapsed and remaining time for the current stage;
     (ii) passively, when the deadline for a stage is approaching and required deliverables are still missing, \method injects a warning message asking the agent to stop exploration and generate the necessary artifacts.
     For API cost, \method tracks accumulated token usage across sessions, but does not expose token consumption information to the agent.
     When the cost limit is reached, the run is aborted and the current workspace is preserved as the final snapshot.
     Budget control also supports operational continuity for long-running research processes.
     \method persists each stage's session identifier, status, elapsed time, and effective budget, so an interrupted run can resume from the latest filesystem state under the remaining budget rather than restarting from scratch.
     Users may also revise the configured time limits, or grant explicit extra resume time when a stage has exhausted its budget before producing required artifacts.
     This makes budget engineering not only a stopping rule, but also an operational interface for controlled continuation.

     \begin{figure}[!t]
      \centering
      \includegraphics[width=\linewidth]{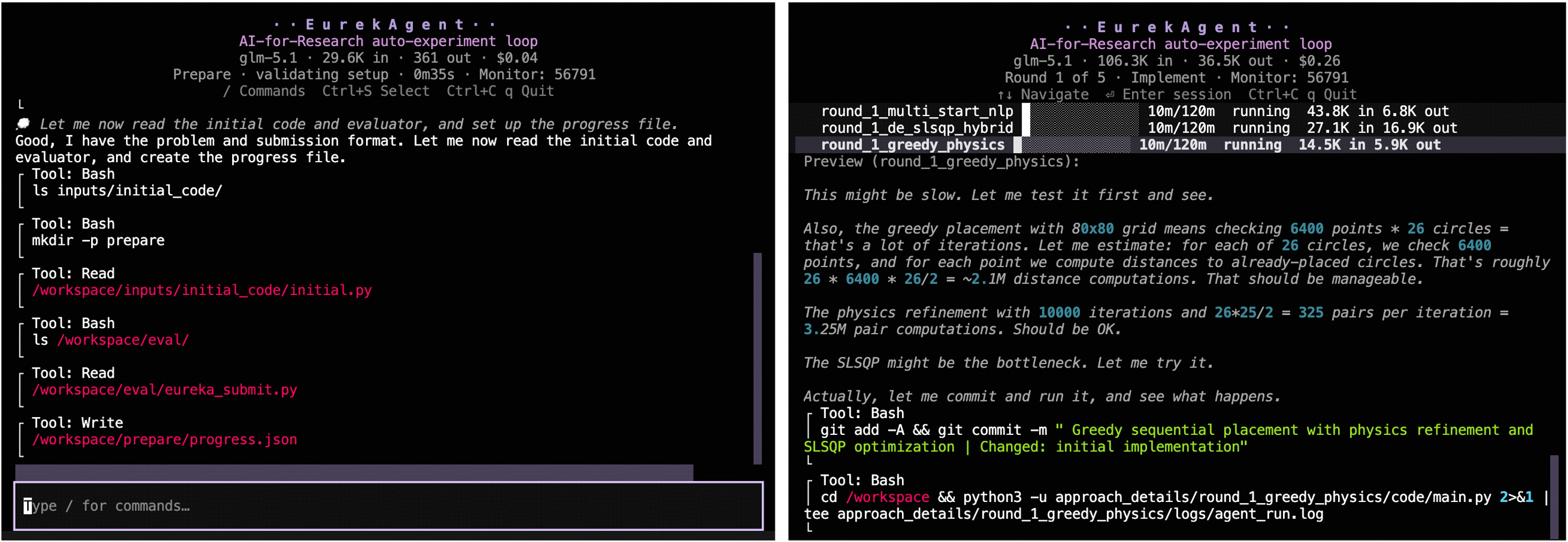}
      \caption{The \method terminal UI interface. The terminal UI preserves a CLI-agent style view for inspecting live session outputs and communicating with agents through the bottom input box. Left: prepare-stage
      snapshot. Right: implement-stage snapshot with three parallel implementation sessions; users can preview all sessions or enter any session to inspect details and communicate.}
      \label{fig:tui}
      \end{figure}
   
   \paragraph{Human-in-the-loop Engineering.}
   Although \method supports fully autonomous iteration, scientific discovery still benefits from human oversight.
   %  and timely intervention.
   \method therefore provides two complementary interfaces.
   On one hand, a terminal UI (Figure~\ref{fig:tui}) exposes per-approach progress, raw session outputs, and an input box for users to communicate with active sessions.
   On the other hand, a web monitor (Figure~\ref{fig:web-monitor}) provides higher-level views of the run, presenting a visualized score evolution with per-round and global best approaches.
   % For intervention, the user can interrupt an active session from the terminal UI and send guidance, corrections, or comments to the
   % corresponding session.
   % The preparation stage can also pause and request structured clarification when the task setup is ambiguous.
   % These interfaces preserve autonomy as the default mode, while making the process fully observable and allowing humans to redirect agent
   % behavior when needed.
   These interfaces preserve agent autonomy while keeping the process fully observable and allowing humans to redirect agent behavior when needed.

   \section{Experiments}
   
   We evaluate \method on three domains: mathematics, kernel engineering, and machine learning engineering.
   We focus on tasks with optimizable metrics, where progress can be measured by objective scores.
   All experiments use \method
   %  in a training-free setting 
   with Claude Code as the CLI agent and GLM-5.1 as the base LLM.
   Building on this setting, we configure the Web Search Prime MCP\footnote{\url{https://docs.z.ai/devpack/mcp/search-mcp-server}} to enable search engine capabilities, and the Playwright MCP\footnote{\url{https://github.com/microsoft/playwright-mcp}} to enable web browser navigation.

   \subsection{Mathematics}
   
   We evaluate \method on three mathematical optimization problems following prior work
   %  on autonomous scientific discovery
   ~\citep{alphaevolve,thetaevolve,tttdiscover}:
   (1) circle packing, where the objective is to place 26 non-overlapping circles inside a unit square and maximize the sum of their radii, using the OpenEvolve-style evaluator with a \(10^{-6}\) tolerance for boundary and overlap checks;
   (2) Erd\H{o}s' minimum overlap, where the objective is to minimize the limiting maximum overlap between two equal-size sets; and
   (3) the first autocorrelation inequality, where the objective is to find a nonnegative construction that certifies the tightest known upper bound on an autoconvolution constant.
   For circle packing, we compare against the previous best AI result reported under the same tolerance setting.
   All three problems have verifiable and optimizable objective functions, making them suitable for agentic solution iteration.
   We report \method hyperparameter settings in Appendix~\ref{app:hyperparameters}.
   
   \begin{table}[!h]
     \centering
     \footnotesize
     \setlength{\tabcolsep}{4pt}
     \renewcommand{\arraystretch}{1}
     % \resizebox{\linewidth}{!}{%
     \begin{tabular}{@{}l|cc@{\hspace{0.9em}}|@{\hspace{0.9em}}cc@{}}
     \toprule
     Task & \method & LLM & Prev. Best AI & LLM \\
     \midrule
     Circle Packing ($\uparrow$) &
     \eurekascore{2.635}{999} &
     GLM-5.1 &
     \score{2.635986}\tablecite{thetaevolve} &
     R1-Distill-Qwen3-8B \\
     Erd\H{o}s' Min. Overlap ($\downarrow$) &
     \eurekascore{0.3808}{70} &
     GLM-5.1 &
     \score{0.380876}\tablecite{tttdiscover} &
     gpt-oss-120b \\
     1st Autocorr. Ineq. ($\downarrow$) &
     \eurekascore{1.5028}{61} &
     GLM-5.1 &
     \score{1.502863}\tablecite{tttdiscover} &
     gpt-oss-120b \\
     \bottomrule
     \end{tabular}%
     % }
     \caption{Performance of \method on three mathematical optimization problems. Previous best AI results are from test-time training systems, while \method remains training-free with only environment engineering designs.}
     \label{tab:math-results}
     \end{table}
   
   As shown in Table~\ref{tab:math-results}, \method establishes new state-of-the-art results on all three mathematics tasks.
   Notably, \method outperforms prior test-time training systems while remaining training-free, suggesting that environment engineering alone can unlock breakthrough results
   % autonomous
   % optimization 
   without updating the backbone model.

   \subsection{Kernel Engineering}
   
   We evaluate \method on the GPUMODE TriMul competition, which targets optimized implementations of triangular matrix multiplication. 
   % , a core primitive in AlphaFold-style protein structure models~\citep{tttdiscover}.
   % The task requires implementing the outgoing TriMul forward pass and minimizing runtime while passing correctness tests.
   Submissions are evaluated by the geometric mean runtime across benchmark cases, where lower runtime is better.
   We evaluate on the A100 setting.
   
   Because the official GPUMODE leaderboard closed,
   % on May 9, 2026, 
   we could not submit new solutions and get official scores.
   We therefore evaluate locally on an A100 GPU using the released TTT-Discover TriMul setting~\citep{tttdiscover}, with only minimal format adaptation for \method submissions.
   For fair comparison, we download top leaderboard solution scripts from GPUMODE and regrade them under the same local protocol.
   All candidates are evaluated on the same A100 GPU with the original correctness tests, benchmark cases, scoring rule, and timing logic unchanged.
   We run three warmup rounds, followed by ten measured rounds with randomly shuffled candidate order to reduce order effects, and report both median and mean geometric-mean runtimes.
   For \method, we report the four best solutions discovered throughout a single system run; hyperparameter settings are listed in Appendix~\ref{app:hyperparameters}.
   
   \begin{table}[!h]
     \centering
     \footnotesize
     \setlength{\tabcolsep}{5pt}
     \renewcommand{\arraystretch}{1.12}
     % \resizebox{\linewidth}{!}{%
     \begin{tabular}{@{}cllcc@{}}
     \toprule
     \textbf{Rank} & \textbf{Solution} & \textbf{LLM} & \textbf{Median ($\mu\mathrm{s}$) ($\downarrow$)} & \textbf{Mean ($\mu\mathrm{s}$) ($\downarrow$)} \\
     \midrule
     1 & \method-CUDA Graph & GLM-5.1 & \textbf{2005.0307} & \textbf{2014.1874} \\
     2 & \method-INT8 BMM & GLM-5.1 & \textbf{2006.9998} & \textbf{2013.5141} \\
     3 & \method-Fused Front-End & GLM-5.1 & \textbf{2016.5718} & \textbf{2020.2674} \\
     4 & \method-Triton Autotune & GLM-5.1 & \textbf{2030.6877} & \textbf{2041.5578} \\
     5 & josusamartin & N/A & 2096.0441 & 2105.1655 \\
     6 & TTT-Discover\tablecite{tttdiscover} & gpt-oss-120b & 2247.7849 & 2248.2307 \\
     7 & rd9000 & N/A & 2300.4883 & 2307.5716 \\
     \bottomrule
     \end{tabular}%
     % }
     \caption{Performance of \method on the TriMul kernel engineering task.}
     \label{tab:kernel-results}
     \end{table}
   
   Table~\ref{tab:kernel-results} shows that \method discovers multiple solutions that outperform the top leaderboard submissions under the same local evaluator.
   The top four \method solutions all achieve median runtimes below \(2031\,\mu\mathrm{s}\), indicating stable high-quality optimization rather than a single lucky candidate.
   The best \method kernel improves over the strongest regraded leaderboard solution by about $4.3\%$ and over TTT-Discover by about $10.8\%$.

   \subsection{Machine Learning Engineering}
   
   We evaluate \method on a curated subset of seven competitions from the MLE-Bench Lite split~\citep{mlebench}.
   MLE-Bench evaluates agents on real Kaggle-style machine learning competitions, where submissions are scored against held-out test sets and mapped to medal thresholds.
   To balance cost, diversity, and difficulty, we start from the 22 Lite competitions and use public MLE-Bench leaderboard results to estimate tractability.
   We divide tasks into Easy, Medium, and Hard tiers by aggregate prior-agent medal rate, then sample 2 Easy, 2 Medium, and 3 Hard competitions.
   The selected competitions span image, text, audio, and tabular prediction.
   Our selected tasks are detailed in Appendix~\ref{app:mlebench-subset}.
   
   We run \method once per competition and report the resulting medal rates.
   Following MLE-Bench's official 24-hour and single-GPU setting, we grant one GPU to each run and report \method hyperparameter settings in Appendix~\ref{app:hyperparameters}.
   For baselines, we use the corresponding public MLE-Bench leaderboard results on the same tasks.
   When a baseline reports the aggregated scores of multiple runs, we report the upper end of its reported score range.
   
   \begin{table}[!h]
     \centering
     \footnotesize
     \setlength{\tabcolsep}{5pt}
     \renewcommand{\arraystretch}{1.12}
     % \resizebox{\linewidth}{!}{%
     \begin{tabular}{@{}cllccc@{}}
     \toprule
     \textbf{Rank} & \textbf{Agent} & \textbf{LLM} & \textbf{Any Medal} & \textbf{Gold} & \textbf{Above Median} \\
     \midrule
     \multicolumn{1}{c}{1} & \method & GLM-5.1 & \textbf{85.71\%} & \textbf{71.43\%} & \textbf{100.00\%} \\
     \multicolumn{1}{c}{2} & AIBuildAI\tablecite{aibuildai} & Claude-Opus-4.6 & \underline{71.43\%} & 57.14\% & 85.71\% \\
    \multicolumn{1}{c}{3} & Famou-Agent\tablecite{li2025fmagent} & Gemini-2.5-Pro & \underline{71.43\%} & 57.14\% & \textbf{100.00\%} \\
    \multicolumn{1}{c}{4} & Famou-Agent 2.0\tablecite{li2025fmagent} & Gemini-2.5-Pro & \underline{71.43\%} & \underline{65.39\%} & \underline{95.24\%} \\
     \multicolumn{1}{c}{5} & LoongFlow\tablecite{loongflow} & Gemini-3-Flash-Preview & \underline{71.43\%} & 57.14\% & 71.43\% \\
     \multicolumn{1}{c}{6} & CAIR MARS+\tablecite{cairmars} & Gemini-3-Pro-Preview & \underline{71.43\%} & \textbf{71.43\%} & \textbf{100.00\%} \\
     \bottomrule
     \end{tabular}%
     % }
     \caption{Machine learning engineering results on our seven-task MLE-Bench Lite subset. For baselines with multiple runs, we report the upper end of the reported range.}
     \label{tab:mle-results}
     \end{table}
   
   As shown in Table~\ref{tab:mle-results}, \method achieves the highest any-medal rate on the selected MLE-Bench subset, reaching $85.71\%$ with a single run per task.
   It also attains the highest gold-medal rate among methods using non-commercial open models.
   All listed baselines use closed commercial models, while \method runs with open-source LLM GLM-5.1, suggesting that environment-engineered autonomous iteration can be competitive even without
   relying on the strongest proprietary models.
   
   \section{Conclusion and Limitations}
   
   % Summarize \method
   
   % While we make every effort to ensure the reliability and easy usability of \method, as scientific research is a complex and dynamic process, it is possible that \method may still contain flaws in some cases. 
   % We will continue to develop \method, and sincerely welcome contributions to improve it.
   
   We presented \method, an environment-engineered system for autonomous scientific discovery on metric-driven research tasks. Rather than prescribing detailed research
     workflows, \method coordinates off-the-shelf CLI-agent sessions through a simple prepare-propose-implement loop, while shaping the surrounding environment for reliable
     evaluation, shared progress memory, resource boundaries, and human oversight.
     Using Claude Code as the CLI agent and GLM-5.1 as the base LLM, \method achieves new state-of-the-art results on all evaluated mathematics and kernel engineering tasks, and
     ranks first on our evaluated MLE-Bench Lite subset. These results suggest that, as general-purpose CLI agents become more capable, carefully engineered scientific-discovery
     environments can turn model capability into reliable scientific progress.
   
   Looking forward, we view environment engineering as a central layer in the next generation of autonomous research systems.
   As agents become more capable, scientific progress will depend not only on model intelligence, but also on the environments that define reliable feedback, persistent memory, resource control, evaluator integrity, human oversight, and recoverable long-running operation.
   Our current experiments focus on metric-driven tasks with executable evaluators, but the same perspective becomes even more important as autonomous research moves toward broader and more open-ended scientific settings.
   
   We open-source \method as an initial step toward this direction and invite the community to build on, improve, and contribute to it.
   We will continue maintaining the project, extending it to richer research domains, and updating empirical results on its performance, capabilities, and boundaries.
   We hope \method can serve as a practical starting point for collective exploration of environment engineering as a foundation for reliable autonomous scientific discovery.

\newpage

\bibliography{iclr2026_conference}
\bibliographystyle{iclr2026_conference}

\newpage
\appendix
\section{\method Hyperparameter Settings}
\label{app:hyperparameters}

Table~\ref{tab:hyperparameters} summarizes the \method hyperparameters used in our experiments.
Here, \(R\) denotes the maximum number of propose--implement iteration rounds, and \(P\) denotes the maximum number of parallel implementation sessions spawned in each implement stage.

\begin{table}[!h]
  \centering
  \footnotesize
  \setlength{\tabcolsep}{5pt}
  \renewcommand{\arraystretch}{1.12}
  \begin{tabular}{@{}lccccp{0.24\linewidth}@{}}
  \toprule
  \textbf{Task} & \(\boldsymbol{R}\) & \(\boldsymbol{P}\) & \(\boldsymbol{t_{propose}}\) & \(\boldsymbol{t_{implement}}\) & \textbf{Notes} \\
  \midrule
  Circle Packing & 5 & 3 & 20 min & 120 min & -- \\
  Erd\H{o}s' Min. Overlap & 8 & 3 & 20 min & 120 min & -- \\
  1st Autocorr. Ineq. & 8 & 3 & 20 min & 120 min & -- \\
  TriMul & 13 & 3 & 20 min & 160 min & A100 evaluation setting. \\
  MLE-Bench Lite & 12 & 3 & 20 min & 100 min & One GPU per run. \\
  \bottomrule
  \end{tabular}
  \caption{\method hyperparameter settings used in our experiments.}
  \label{tab:hyperparameters}
  \end{table}

\section{Selected MLE-Bench Lite Competitions}
\label{app:mlebench-subset}

We select seven MLE-Bench Lite competitions across three difficulty tiers, using aggregate medal rates of prior public leaderboard agents as a proxy for task difficulty:
\begin{itemize}
  \item \textbf{Easy} ($>40\%$): histopathologic-cancer-detection (57.0\%) and plant-pathology-2020-fgvc7 (49.7\%).
  \item \textbf{Medium} ($15\%$--$40\%$): aerial-cactus-identification (26.2\%) and the-icml-2013-whale-redux (23.5\%).
  \item \textbf{Hard} ($<10\%$): jigsaw-toxic-comment (9.1\%), dog-breed-identification (0.4\%), and tabular-playground-may-2022 (0.4\%).
\end{itemize}

\end{document}